\documentclass[sn-mathphys]{JORSC-jnl}

\jyear{2023}%

\theoremstyle{thmstyleone}%
\theoremstyle{thmstylethree}%
\newtheorem{definition}{Definition}%

\usepackage{amsmath}
\usepackage{indentfirst}
\usepackage{epstopdf}
\begin{document}
\renewcommand{\qedsymbol}{}
\title[Recent Advances in Hypergraph
Neural Networks]{Recent Advances in Hypergraph
Neural Networks}


\author[1]{Murong Yang}\email{mryang@shu.edu.cn}

\author*[2]{Xin-Jian Xu}\email{xinjxu@shu.edu.cn}

\affil[1]{\orgdiv{Department of Mathematics,  College of Sciences}, \orgname{Shanghai University}, \orgaddress{ \city{Shanghai} \postcode{200444}, \country{China}}}

\affil[2]{\orgdiv{Qian Weichang College}, \orgname{Shanghai University}, \orgaddress{ \city{Shanghai University} \postcode{200444}, \country{China}}}

\equalcont{This work was supported by
the National Natural Science Foundation of China (No.12071281)}


\abstract{
The growing interest in hypergraph neural networks (HGNNs) is driven by their capacity to 
capture the complex relationships and patterns within hypergraph structured data across various domains,  including computer vision, complex networks, and natural language processing. This paper comprehensively reviews  recent advances in HGNNs and 
presents a taxonomy of mainstream models based on their  architectures:
hypergraph convolutional networks (HGCNs), hypergraph attention networks (HGATs), hypergraph autoencoders (HGAEs), hypergraph recurrent networks (HGRNs), and deep hypergraph generative models (DHGGMs).
For each category, we delve into its practical applications, mathematical mechanisms, literature contributions, and open problems.
Finally, we discuss 
some common challenges and promising research directions.
This paper aspires to be a helpful resource that provides guidance for future research and applications of HGNNs.} 




\maketitle

\section{Introduction}\label{sec1}

Graph neural networks (GNNs) have been rapidly developed for their expressive power, flexible modeling, and end-to-end training capabilities. They are applied in various domains, such as drug discovery, image processing, and traffic prediction \cite{2018Drug,20173D,wordgcn2019,2018Spatio}.
However, real-world data often involves higher-order relationships.
For instance, in academic collaboration networks, a single paper may be co-authored by multiple researchers; in social networks, an event may involve interactions among multiple group members; and in bioinformatics, a biological process may be regulated by the coordinated action of multiple genes. These relationships exceed the representational capacity of simple graphs.
To better capture this complexity,
hypergraphs are used to describe these higher-order relationships by connecting more than two nodes. 
Thus, it is essential to develop effective hypergraph learning models.
As a mainstream method, hypergraph neural networks (HGNNs) extend the capabilities of neural networks to flexibly model and analyze such complex data. 
This flexibility has attracted growing research interest and has been widely applied across various fields, 
such as computer vision \cite{yu2012adaptive,2022Hypergraph}, complex networks \cite{9679118,10.1145/3442381.3449844}, and natural language
processing \cite{ding2020less,Malik2023}.

In the research of HGNNs, scholars have proposed various architectures:

Hypergraph convolutional networks are an important type that represent hypergraph by performing convolution operations directly on it.
Feng et al. \cite{2019Hypergraph} introduce the first spectral hypergraph convolution  based on Zhou's normalized hypergraph Laplacian.
Meanwhile, Fu et al. \cite{2019HpLapGCN} propose the HpLapGCN model with the hypergraph p-Laplacian. Later, Nong et al. \cite{wavelet} introduce the HGWNN model with wavelet localization instead of Fourier transform.
Furthermore, models like \cite{2018HyperGCN} by Yadati et al. transform hypergraphs into simple graphs and apply spectral graph convolution.
During the same period, spatial hypergraph convolution models have also been developed. Particularly,  Arya et al. \cite{arya2020hypersage} design the HyperSAGE model by extending GraphSAGE  \cite{2017Inductive} to hypergraphs. 
To further enhance, Huang et al. \cite{2021UniGNN} propose a unified message-passing framework,
UniGNN. 
Subsequently, Chien et al. \cite{chien2021you} develpo the AllSet model by  using permutation-invariant set functions for message-passing. 
And Telyatnikov et al. \cite{telyatnikov2023hypergraph} propose the MultiSet model by incorporating multi-set functions for higher-order homogeneous network.

Since hypergraph convolution networks treat all neighbors equally that may limit model performance, scholars have developed hypergraph attention networks to address this issue.
Zhang et al. \cite{2020Hyper} propose a self-attention-based model (Hyper-SAGNN) for homogeneous and heterogeneous hypergraphs. Later, Ding et al. \cite{ding2020less} develop a hypergraph attention network for text classification with node-level and hyperedge-level attention. Kim et al. \cite{kim2020hypergraph} introduce a co-attention model to address information disparity in multimodal learning. Additionally, Chen et al. \cite{Chen2020Hypergraph} propose a hypergraph attention network that dynamically assigns weights to nodes and hyperedges. Afterward, Bai et al. \cite{bai2021hypergraph} introduce hypergraph convolutional and attention operators (HCHA) for node classification, and Wang et al. \cite{Wang2021Session-based} develop a hypergraph attention-based system for dynamic item embedding generation.


For unsupervised learning, hypergraph autoencoders have been developed.
Early studies follow the traditional autoencoder by using features as the sole input with hypergraph Laplacian regularization. 
Hong et al. \cite{hong2016hypergraph} first introduce the HRA model that combines denoising autoencoders with this regularization for local preservation. Another approach includes both features and structure as inputs. 
Banka et al. \cite{banka2020multi} develop a hyperconnectome autoencoder with hyperconnected groups and hypergraph convolutional layers to improve brain connectivity mapping. Hu et al. \cite{hu2021adaptive} design an adaptive hypergraph autoencoder (AHGAE) with a smoothing filter to reduce noise and enhance features. Later, Shi et al. \cite{SHI2023110172} propose a fault diagnosis method that employs hypergraphs to analyze unlabelled vibration signals and leverages deep hypergraph autoencoder embeddings.


To process
sequential hypergraph data, hypergraph recurrent networks have also been developed. 
As an early exploration, Yi et al. \cite{2020HGC-RNN} first combine hypergraph convolution with recurrent neural networks to capture structural and temporal dependencies. In subsequent studies, Yu et al. \cite{2023RHCRN} develop a routing hypergraph convolutional recurrent network (RHCRN) for traffic prediction by integrating gated recurrent units for spatiotemporal correlations. Recently, Yang et al. \cite{2024HRNN} propose a hypergraph recurrent neural network  
to capture temporal features and
integrate spatiotemporal semantics for intrusion detection.


Besides the models mentioned above, research on deep hypergraph generative models has also gained traction 
with distinct approaches.
Early model incorporates hypergraph constraints into the latent space of VAEs, as exemplified by Kajino's molecular hypergraph grammar variational autoencoder (MHG-VAE) \cite{kajino2019molecular}. Later, Fan et al. \cite{2021Heterogeneous} incorporate the hypergraph structure into the input of encoder and propose a heterogeneous hypergraph variational autoencoder (HeteHG-VAE) for link prediction.
Liu et al. \cite{liu2022hypergraph} develop a hypergraph variational autoencoder for high-order interactions in multimodal data. Based on the generative adversarial mechanism, Pan et al. \cite{Pan2021HGGAN} propose the HGGAN model for Alzheimer's analysis.  And Bi et al. \cite{Bi2022IHGC-GAN} create an influence hypergraph convolutional generative adversarial network (IHGC-GAN) for disease risk prediction.
Most recently, based on the hypergraph diffusion process, Gailhard et al. \cite{gailhard2024hygene} propose the HYGENE model that progressively reconstructs global structures and local details through a denoising process.



Currently,
Wang et al. \cite{wang2022survey} provide a comprehensive survey of various HGNN models solely for node classification in the context of action recognition.
And Kim et al. \cite{kim2024survey} conduct a detailed analysis of the internal components within HGNN models.
However, there is still a lack of systematic exploration
and categorization of HGNNs.
In this paper, we review  recent advances in HGNNs and categorize  their architectures into five types: hypergraph convolutional networks (HGCNs), hypergraph attention
networks (HGATs), hypergraph autoencoders (HGAEs),  hypergraph recurrent
networks (HGRNs),  and deep
hypergraph generative models (DHGGMs). For each category, we conduct an in-depth analysis in terms of practical applications, mathematical mechanisms, research contributions, and open problems.

The rest of this paper is organized as follows. Section \ref{sec2} introduces the definitions of hypergraphs and  tasks of HGNNs.
Section \ref{sec3} provides a taxonomy of HGNN architectures along with a comprehensive overview. Section \ref{sec4} concludes the paper and outlines potential research directions for HGNNs.


\section{Preliminaries}\label{sec2}

\subsection{Definitions of Hypergraphs}
We review the definitions of some basic hypergraphs, including undirected, directed, heterogeneous, and temporal hypergraphs.
\vspace{-3mm}
\begin{definition}\textbf{[Undirected Hypergraph]}
An undirected hypergraph is represented by $\mathcal{G}=(\mathcal{V}, \mathcal{E})$, where $\mathcal{V}$ is a finite set of nodes and $\mathcal{E}$ is a finite set of hyperedges.
If each node has a corresponding $d$-dimensional feature or attribute vector, they are stacked into a feature matrix $\mathbf{X} \in \mathbb{R}^{\mid\mathcal{V}\mid \times d}$.
The relationships between nodes and hyperedges are represented by the incidence matrix $\mathbf{H}\in \mathbb{R}^{\mid\mathcal{V}\mid \times\mid\mathcal{E}\mid}$, where the elements of the incidence matrix $\mathbf{H}$ are defined as:
$$
\mathbf{H}(v, e)=\left\{\begin{array}{l}
1, \,\,\,\, v \in e \\
0, \,\,\,\, v \notin e
\end{array}\right.
$$

In a weighted undirected hypergraph,  $\mathcal{G}=(\mathcal{V}, \mathcal{E}, \mathbf{W})$, where the diagonal matrix $\mathbf{W}$ is the hyperedge weight matrix, and the diagonal elements represent the weight of each hyperedge $e$, denoted as $w(e)$.
Furthermore, the degree of hyperedge $e$ is defined as the number of nodes contained in the hyperedge, i.e.,
$\delta(e)=\sum\limits_{v \in \mathcal{V}} \mathbf{H}(v, e)$, and the degree of node $v$ is defined as the sum of the weights of all hyperedges containing that node, i.e.,
$d(v)=\sum_{e \in \mathcal{E}} w(e) \mathbf{H}(v, e)$.
Assuming $\mathbf{D}_e$ and $\mathbf{D}_v$ represent the diagonal matrices of hyperedge degrees and node degrees, respectively.
The normalized hypergraph Laplacian matrix is defined as $
\Delta=\mathbf{I}-\mathbf{D}_v^{-1 / 2} \mathbf{H} \mathbf{W} \mathbf{D}_e^{-1} \mathbf{H}^T \mathbf{D}_v^{-1 / 2}
$, where $\mathbf{I}$ is the identity matrix.
\end{definition}

\vspace{-9mm}
\begin{definition}\textbf{[Directed Hypergraph]}
A directed hypergraph is represented by $\overrightarrow{\mathcal{G}}=(\mathcal{V}, \overrightarrow{\mathcal{E}})$, where $\mathcal{V}$ is a finite set of nodes and $\overrightarrow{\mathcal{E}}$ is a finite set of hyperarcs. $\mid\mathcal{V}\mid$ and $\mid\overrightarrow{\mathcal{E}}\mid$ denote the number of nodes and hyperarcs, respectively. Each hyperarc $\overrightarrow{e}=(\overrightarrow{e^{+}}, \overrightarrow{e^{-}}) \in \overrightarrow{\mathcal{E}}$, where $\overrightarrow{e_i^{+}} \in \mathcal{V}$ represents the tail (source node) of hyperarc $\overrightarrow{e_i}$, and $\overrightarrow{e_i^{-}} \in \mathcal{V}$ represents the head (target node) of hyperarc $\overrightarrow{e_i}$. It holds that $\overrightarrow{e_i^{+}} \cap \overrightarrow{e_i^{-}}=\varnothing$, $\overrightarrow{e_i^{+}} \neq \varnothing$, and $\overrightarrow{e_i^{-}} \neq \varnothing$. 
The incidence matrix is trivially defined as

\begin{equation}
\overrightarrow{\mathbf{H}}(v, e)=\left\{\begin{array}{c}
-1 , \, \text{if } v = \overrightarrow{e^{+}} \\
1 , \, \text{if } v = \overrightarrow{e^{-}} \\
0 , \, \text{otherwise}
\end{array}\right.
\end{equation}

The incidence matrix $\overrightarrow{\mathbf{H}}$ can be considered as composed of two different matrices $\mathbf{H}^+$ and $\mathbf{H}^-$, which describe the association between each hyperedge and the source and target nodes of that hyperedge, respectively. These two incidence matrices guide the message passing on the directed hypergraph. The degree matrix of source nodes $\mathbf{D}^{+}$ and the degree matrix of target nodes $\mathbf{D}^{-}$ correspond to the diagonal matrices of the sum of the columns of $\mathbf{H}^+$ and $\mathbf{H}^-$, respectively.
\end{definition}

\vspace{-9mm}
\begin{definition}\textbf{[Heterogeneous Hypergraph]}
A heterogeneous hypergraph is a more general and flexible form of hypergraph that allows different types of nodes to be connected together to form hyperedges. A heterogeneous hypergraph is represented as $\mathcal{G}=(\mathcal{V}, \mathcal{E})$, where the node set $\mathcal{V}=\bigcup_{i=1}^n \mathcal{V}_i$ is composed of a union of different types of node sets, and $\mathcal{E}$ is a finite set of hyperedges. If there exists a hyperedge $e\in\mathcal{E}$ that connects nodes from different types of node sets $\mathcal{V}_{i_1}, \mathcal{V}_{i_2}, \ldots, \mathcal{V}_{i_k}$, then hyperedge $e$ is a heterogeneous hyperedge, and \( \mathcal{G }\) is a heterogeneous hypergraph.
\end{definition}

\vspace{-9mm}
\begin{definition}\textbf{[Temporal Hypergraph]}
A temporal hypergraph incorporates a temporal dimension compared to the traditional hypergraph. It contain a time-series of static hypergraphs $\mathcal{G}^{(\mathrm{t})}=(\mathcal{V}^{(\mathrm{t})}, \mathcal{E}^{(\mathrm{t})})$, where $\mathrm{t} = 1, \ldots, \mathrm{T} $.
If $\mathcal{V}^{(\mathrm{1})}= \mathcal{V}^{(\mathrm{2})}=\cdots=\mathcal{V}^{(\mathrm{T})}$ indicates that the node set does not change over time. Correspondingly, a incidence matrix sequence $\left\{\mathbf{H}^{(\mathrm{1})}, \mathbf{H}^{(\mathrm{2})}, \ldots, \mathbf{H}^{(\mathrm{T})}\right\}$ can be defined to represent the temporal evolution of high-order associations between nodes.
\end{definition}

%

\subsection{Tasks of HGNNs}\label{subsec2}
Based on hierarchical categorization, the tasks of HGNNs on hypergraph data are mainly divided into the following three levels:

\begin{itemize}
   \item[i)] \textbf{Node-level} tasks primarily include node classification or node clustering.
Node classification aims to predict the labels of nodes, such as classifying documents based on their keyword sets or topics in document analysis. Node clustering is to group nodes into different clusters, such as community detection in social networks.
\vspace{1mm}

  \item[ii)] \textbf{Hyperedge-level} tasks primarily include  hyperedge classification or hyperedge prediction. Hyperedge classification aims to predict the labels of hyperedges, such as predicting the thematic categories of research collaboration papers in academic networks. Hyperedge prediction aims to predict which set of nodes may form a new hyperedge. For example, it can be used to predict whether different chemical groups may combine to form new molecular structures in drug discovery.
\vspace{1mm}

    \item[iii)] \textbf{Hypergraph-level} tasks primarily include hypergraph classification or hypergraph generation. Hypergraph classification aims to predict the labels of the entire hypergraph, such as community classification in social networks.
        Hypergraph generation aims to generate new hypergraph structures or node features. For example, generating images with specific features, such as synthetic human faces or medical images.
\end{itemize}


Some tasks cannot be directly classified into the above three categories, especially derivative tasks, such as analyzing temporal hypergraphs. 
Nevertheless, these tasks are typically based on the representation at a specific level of the hypergraph as well.

\section{Hypergraph Neural Networks}\label{sec3}

\begin{figure}[htbp]
  \centering
  \includegraphics[width=0.99\linewidth]{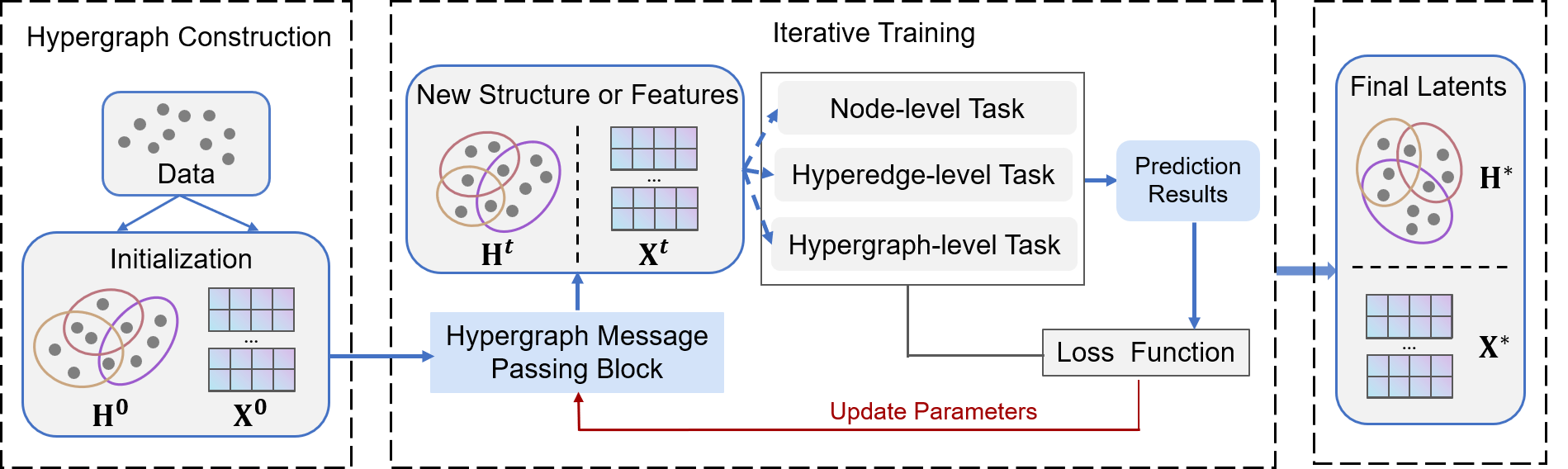}\\
  \caption{The overall framework of HGNNs.
  The purpose of the hypergraph message passing block is to update the hypergraph structure $\mathbf{H}$ or features $\mathbf{X}$. It may involve various techniques, such as hypergraph convolution in HGCNs, attention-weighted aggregation in HGATs, encoding and decoding in HGAEs, recursive message passing in HGRNs, or hypergraph generation in DHGGMs.}
  \label{fig1}
\end{figure}

In this section, we introduce our taxonomy of HGNNs and divide them into HGCNs, HGATs, HGAEs, HGRNs, and DHGGMs.
We present  a general framework for HGNNs, as illustrated in Fig. \ref{fig1}. It begins with hypergraph construction to initialize the hypergraph structure and features. Next, the hypergraph message passing block aggregates information from neighbors to update the hypergraph structure, node features, or hyperedge features (determined by the task). Then, task predictions are made based on the updated structure or features. Following this, the loss is computed to update model parameters  through backpropagation. This process is iteratively refined to optimize the latent structure or features, thereby enhancing the model's performance on the task.

\subsection{Hypergraph Convolutional Networks}
Hypergraph convolutional networks (HGCNs) are among the most mainstream models in HGNNs.
HGCNs learn feature representations of nodes or hyperedges through hypergraph convolution operations. They are typically applied in supervised, or weakly supervised tasks.
They demonstrate strong potential in various fields, including hyperspectral image classification \cite{9590574}, group and social recommendation \cite{9679118,10.1145/3442381.3449844}, session-based recommendation \cite{Xia_Yin_Yu_Wang_Cui_Zhang_2021}, and weakly supervised semantic segmentation \cite{9897774}.
In this subsection, we review HGCNs, which are categorized into spectral HGCNs and spatial HGCNs.
\subsubsection{Spectral HGCNs}


%

The common approach for spectral HGCNs is to define convolution operations in the spectral domain of hypergraphs 
through hypergraph Laplacian eigen-decomposition.  
Similar to spectral GCNs \cite{NIPS2016_04df4d43}, the key mathematical form of spectral HGCNs is:
\begin{equation}
\label{eq2}
\mathbf{x} *_\mathcal{G } \mathbf{y}
=\mathbf{U} g_\theta(\Lambda) \mathbf{U}^T \mathbf{x}
\end{equation}
where $\mathbf{x} *_\mathcal{G } \mathbf{y}$
represents the convolution of the node signal
\( \mathbf{x}\) by using the filter \( \mathbf{y}
\) on the hypergraph
$\mathcal{G}$. $\mathbf{U}$ is the eigenvector matrix of the hypergraph Laplacian, and $\Lambda$ is the eigenvalue matrix. $g_\theta(\Lambda)$ represents a learnable, parameterized filter function defined on $\Lambda$. It could be a polynomial function of $\Lambda$ (such as Chebyshev polynomials) or other basis function forms.

As the first proposed spectral HGCN model based on formula (\ref{eq2}), Feng et al. \cite{2019Hypergraph} approximate the hypergraph convolution operation by using the hypergraph Laplacian operator and Chebyshev polynomial. Similar to GCN, the convolution operation is defined by Zhou's normalized
Laplacian matrix \cite{zhou2006learning}. Meanwhile, Fu et al.  \cite{2019HpLapGCN} propose the HpLapGCN model with the hypergraph p-Laplacian, which derives the first-order approximation of spectral hypergraph p-Laplacian convolution. This model provides a more effective hierarchical aggregation method for semi-supervised classification. For attributed graph datasets, Wu et al. \cite{2021Dual} construct dual hypergraphs based on the topological structure and the attributes of nodes. They propose the dual-view hypergraph neural networks. And it deeply models and integrates the complex relationships between nodes through shared and specific hypergraph convolutional layers and attention mechanisms. Unlike \cite{2019Hypergraph} that utilizes Fourier transform,
Nong et al. \cite{wavelet} propose a hypergraph wavelet neural network (HGWNN) model. This model leverages the localization properties of wavelets to achieve efficient information aggregation. Later, some variant models of HGNN \cite{2019Hypergraph} emerged. For temporal link prediction in dynamic networks, Huang et al. \cite{HyperDNE} propose an enhanced hypergraph neural network framework. 
It fully utilizes high-order information and the dynamic consistency of the network. For heterogeneous information networks, Li et al.  \cite{2023Hypergraph} introduce an end-to-end hypergraph transformation neural network. It leverages the communication capability between different types of nodes and hyperedges to learn high-order relationships and discover semantic information.
Similarly, Hu et al. \cite{hu2023general} propose a multi-channel hypergraph convolution framework for non-uniform heterogeneous networks.  This framework first transforms arbitrary heterogeneous hypergraphs into bipartite hypergraphs. It then designs a multi-channel attentive hypergraph convolution module to fuse information from different types of nodes.
For multi-label image classification, Wu et al.  \cite{2020AdaHGNN} develop a high-order semantic learning model based on an adaptive hypergraph neural network. This model automatically constructs adaptive hypergraphs using label embeddings, which offers more flexibility and efficiency compared to manually constructed hypergraphs.

Another approach for spectral HGCNs
involves transforming hypergraphs into simple graphs and defining hypergraph convolution operations by using the graph Laplacian matrix.
The key step for this approach is how to convert the hypergraph into a simple graph with minimal information loss.
Based on this,
Yadati et al. \cite{2018HyperGCN} 
use mediator nodes to transform hypergraphs into simple graphs within spectral graph convolution framework. Due to this expansion way,  
the model results are less affected by noisy nodes. Tan et al. \cite{2020Hypergraph} construct adaptive hypergraphs to pre-learn high-order relationships between labels and transform hypergraphs into simple graphs by using clique expansion, subsequently fed into a GCN  framework. Dong et al. \cite{2020HNHN} propose a new hypergraph convolutional network model by applying nonlinear activation functions to nodes and hyperedges and converting hypergraphs into simple graphs. Bandyopadhyay et al. \cite{2020Line} introduce a new technique for applying graph convolution to hypergraphs with varying hyperedge sizes. It transforms hypergraphs into line graphs and utilizes graph convolution. Hayhoe et al. \cite{hayhoe2024transferable} extend the transferability results of graph neural networks to hypergraphs. And they design a convolutional architecture to handle hypergraph-supported signals, which establishes transfer error bounds by using multiple spectrally similar graph representations.


\subsubsection{Spatial HGCNs}

The spatial HGCNs operate directly on the local structure of the hypergraph, which update each node's representation by aggregating information from neighboring nodes. This method avoids the need to reduce the hypergraph to a simple graph and also circumvents complex spectral decomposition. It allows for the flexible design of different feature aggregation and update mechanisms. 
A classic approach is a two-stage message-passing framework that transmits information from nodes to hyperedges and then from hyperedges back to nodes, such as UniGNN \cite{2021UniGNN}:
\begin{equation}
\begin{aligned}
& \textbf{\small stage 1:} \, \,\,\, \mathbf{x}_e=\phi_1(\left\{\mathbf{x}_j\right\}_{j \in e}) \\
& \textbf{\small stage 2:} \, \,\,\, \tilde{\mathbf{x}}_i=\phi_2(\mathbf{x}_i,\left\{\mathbf{x}_e\right\}_{e \in \mathcal{E}_i})
\end{aligned}
\end{equation}
where $\phi_1$ and $\phi_2$ are permutation-invariant functions that aggregate node features and hyperedge features, respectively. And $\mathbf{x}_j$ and $\mathbf{x}_{e}$ represent the feature representations of node $j$ and hyperedge $e$, respectively.
$\mathcal{E}_i$ represents the set of all hyperedges containing node $i$.

As an early exploration, Jiang et al. \cite{2019Dynamic} propose the dynamic hypergraph neural network framework, which consists of two stacked modules: the dynamic hypergraph construction and update module and the hypergraph convolution module, for encoding the high-order relationships.  Yadati  \cite{NEURIPS2020_217eedd1} further introduces a generalized message-passing neural network framework (G-MPNN), then extends it to MPNN-R for handling multi-relational, ordered, and recursive hypergraph structures.
Meanwhile, Arya et al. \cite{arya2020hypersage} extend GraphSAGE \cite{2017Inductive} to hypergraphs and introduce the HyperSAGE model. This model employs a phased message-passing mechanism, first passing messages from nodes to hyperedges and then from hyperedges back to nodes.
However, HyperSAGE still faces challenges in handling the computational complexity caused by varying node neighbors.
To address this issue, Huang et al. \cite{2021UniGNN} propose an unified message-passing framework (UniGNN) for graphs and hypergraphs.  It also defines two aggregation functions for the two-stage message passing process,
and reduces the computational complexity simultaneously. Later, Gao et al. \cite{gao2022hgnn+} extend the original HGNN \cite{2019Hypergraph} to HGNN$^+$ model. This model introduces an innovative adaptive hyperedge fusion strategy to generate an overall hypergraph.
It specifically addresses noisy or false-negative connections in hypergraph structures. To handle the potential noise in hypergraph structures, Cai et al. \cite{cai2022hypergraph} propose the hypergraph structure learning (HSL) framework, which optimizes both the hypergraph structure and HGNN in an end-to-end manner. Further research by Chien et al. \cite{chien2021you} introduced a generalized hypergraph neural network framework (AllSet), proven to be a generalization of MPNN. AllSet uses two multiset functions, which allow the model to adaptively learn the optimal propagation method for the data.
And Heydari et al. \cite{heydari2022message} design a unified message-passing neural network  model, to handle hypergraph data and unify existing hypergraph convolution methods. Telyatnikov et al. \cite{telyatnikov2023hypergraph} introduce the concept of higher-order network homogeneity based on message-passing schemes and develop the hypergraph neural network architecture (MultiSet), which provides new insights for hypergraph representation learning. To further improve message-passing mechanisms, Arya et al. \cite{Arya10612216} propose the HyperMSG model, which employs a two-level neural message-passing strategy to effectively propagate information within and across hyperedges. It captures both local and global node importance by learning attention weights related to node degree centrality. Wang et al. \cite{Wang10671597} propose a traffic prediction framework based on the hypergraph message-passing network (HMSG).  The model employs a two-step message-passing mechanism to efficiently extract both spatial and temporal features of traffic flow, demonstrating excellent performance in intelligent transportation applications.

\subsubsection{Open Problems}
Although HGCNs 
have shown excellent performance in practical tasks,
they also exhibit some limitations:
i) In HGCNs, adding more layers can cause node information loss, referred to as oversmoothing. 
Striking a balance between capturing valuable long-range information and avoiding oversmoothing is crucial \cite{10446277}.
Potential solutions include integrating random walk and personalized PageRank, or introducing initial residual and identity mapping into the model \cite{2022preventingoversmoothing,2022hypergraphequivalency}.
ii) Spectral HGCNs require eigen-decomposition of the hypergraph Laplacian, which is computationally expensive for large-scale structured data \cite{YANG2021117}. Even with the use of Chebyshev polynomials and other approximation techniques to bypass this step, the computation still needs to be carried out on the entire hypergraph, which makes the process highly memory-intensive. 
iii) Spatial HGCNs operate directly on the local structure of hypergraphs, which offers more computational efficiency. 
However, 
the computational complexity is still influenced by the size of nodes and hyperedges. iv) Spatial HGCNs design the flexible aggregation mechanism during the message-passing process.
And the design of effective neighbor sampling 
and the choice of aggregation functions will impact 
the model's performance.

\subsection{Hypergraph Attention Networks}

HGCNs generally assume that all neighbor nodes or hyperedges hold equal importance to the centroid. 
This will lead to reduced performance in tasks that depend on varying neighbor contributions.
Therefore, the attention mechanism is designed to tackle this issue, and these models are known as hypergraph attention networks (HGATs).
They filter out irrelevant information and extract
critical information from data to improve task performance. 
HGATs are widely applied in various fields, including multimodal learning \cite{kim2020hypergraph}, functional brain network classification \cite{JI20221301}, inductive text classification \cite{ding2020less}, social recommendation \cite{Khan2023}, and session-based recommendation \cite{Wang2021Session-based}.

In a common HGAT framework, a shared attention mechanism is typically used to measure the attention coefficients between nodes and their corresponding hyperedges.
\begin{equation}
\alpha_{i j} =a (\mathbf{W} \mathbf{x}_i ; \mathbf{W} \mathbf{x}_{e_j})
\end{equation}
where $a(\cdot)$ denotes a trainable attention function, and $\mathbf{W}$ is a trainable weight matrix. $\mathbf{x}_i$ and $\mathbf{x}_{e_j}$ represent the feature representations of node $i$ and hyperedge $e_j$, respectively. To ensure that the sum of the attention scores for all nodes within the same hyperedge is equal to $1$, the attention scores need to be normalized, typically by using the softmax function.

In existing studies, scholars have proposed various HGATs and their variants. For instance,
Zhang et al. \cite{2020Hyper} propose a self-attention based graph neural network for hypergraphs (Hyper-SAGNN). This model can handle both homogeneous and heterogeneous hypergraphs for network reconstruction.
Ding et al. \cite{ding2020less} develop a hypergraph attention network for text classification, which incorporates a node-level attention and a hyperedge-level attention. 
Meanwhile, Kim et al. \cite{kim2020hypergraph} introduce the hypergraph attention network, which uses co-attention hypergraphs to handle information level disparity in multimodal learning. Chen et al. \cite{Chen2020Hypergraph} propose a hypergraph attention network that dynamically assigns weights to vertices and hyperedges during information propagation. Similarly, Bandyopadhyay et al. \cite{band2020hypergraph} design a model by directly learning edge relationship weights on hypergraphs with a self-attention mechanism. It avoids the cumbersome process of explicitly constructing line graphs. Additionally,
Khan et al. \cite{Khan2023} develop the heterogeneous hypergraph neural network, which captures high-order associations and enhances recommendation accuracy.
In session-based recommendation tasks, Wang et al. \cite{Wang2021Session-based} develop a hypergraph attention-based system for dynamically generating item embeddings. For stock prediction, Sawhney et al. \cite{sawhney2021stock} propose the hypergraph attention model, which combines hypergraph and temporal Hawkes attention mechanisms to optimize stock ranking. 
Tavakoli et al. \cite{tavakoli2022rxn} use hypergraph structures for automatic chemical reaction representation and prediction. And Ji et al. \cite{JI20221301} design the functional connectivity hypergraph attention network for brain disease classification through dynamic hypergraph generation.
Luo et al. \cite{luo2022directed} propose the directed hypergraph attention network (DHAT), which represents road networks using directed hypergraph structures. And it effectively captures spatiotemporal dependencies in traffic data by combining directed hypergraph convolutions.
Ouyang et al. \cite{OUYANG2024123412} introduce the attribute-specific hypergraph attention network to improve sentiment classification with a syntactic distance-based weight allocation mechanism. Bai et al. \cite{bai2021hypergraph} propose hypergraph convolutional and hypergraph  attention operators (HCHA) to achieve significant results for semi-supervised node classification. Georgiev et al. \cite{georgiev2022heat} introduce the HEAT model, which integrates message-passing neural networks and Transformer mechanisms for knowledge base completion and code error detection.
Chai et al. \cite{chai2024hypergraph} propose the multi-view hypergraph learning method, which utilizes node-level and hyperedge-level attention to capture high-order structural information.

\vspace{2mm}
Attention mechanism has become a key approach to enhance the performance of HGNN models. Nonetheless, HGATs face several challenges:
i) 
They inherit the similar limitations of HGCNs, such as high computational complexity, oversmoothing, and challenges in the design of effective neighbor sampling and aggregation.
ii) To handle data with spatial-temporal dependencies, 
the development of more adaptive attention mechanisms is essential.
For instance, in a traffic system, the attention scores of neighboring entities may change dynamically over time and space as the central entity evolves, which makes the undirected hypergraph attention computation less effective \cite{luo2022directed}.
iii) Attention weights assigned to nodes or hyperedges are significantly influenced by the quality of their feature representations and noise in data, ultimately impacting the model's performance.



\subsection{Hypergraph  Autoencoders}

Hypergraph autoencoders (HGAEs) are unsupervised HGNN models.
They reconstruct the hypergraph structure by utilizing the latent representations obtained from encoding the hypergraph structure and node features. In this process, unconnected nodes with similar features in the original hypergraph may form new hyperedges.
HGAEs have been successfully applied to various fields such as image processing \cite{hong2016hypergraph,cai2021hypergraph}, complex networks \cite{li2023shgae}, brain disease classification \cite{banka2020multi,XIE2024107992}, and fault diagnosis \cite{SHI2023110172,10535302}.

Similar to graph autoencoders \cite{9732192}, a hypergraph autoencoder generally consists of an encoder with parameters and a decoder:
\begin{equation}
\label{eq5ed}
\begin{aligned}
& \textbf{\small Encoder:} \,\,\,\,  \mathbf{Z} = \text{HGCN} (\mathbf{X}, \mathbf{H}) \\
& \textbf{\small Decoder:} \,\,\,\, \hat{\mathbf{H}} = \sigma(\mathbf{Z}\mathbf{Z}^{\top})
\end{aligned}
\end{equation}
where 
$\sigma$ represents the Sigmoid function, and $\mathbf{H}$ is the incidence matrix. The encoder maps the node representations to a low-dimensional latent space based on both features $\mathbf{X}$ and structure $\mathbf{H}$. 
Then the decoder is employed to reconstruct the structure. 
The latent representation $\mathbf{Z}$
can be learned by minimizing the reconstruction error between $\hat{\mathbf{H}}$
 and $\mathbf{H}$ (e.g., cross-entropy loss).

In the earliest studies, one approach is to follow the traditional autoencoder framework that uses only features as input, coupled  with the hypergraph Laplacian regularization 
to ensure smoothness constraints.  For instance,
Hong et al. \cite{hong2016hypergraph} present the HRA model 
by combining denoising autoencoders with hypergraph Laplacian regularization to ensure local preservation.
Building on this, Hong et al. \cite{Hong2017} introduce a hypergraph Laplacian matrix based on distance statistics of neighboring pairs to enhance the retention of high-order relationships between samples.
In domain adaptation, Wang et al. \cite{Wang2019denoising} propose a new domain adaptation network, which utilizes denoising autoencoders to extract robust features and explores high-order relationships through hypergraph regularization.
In hyperspectral image processing, Cai et al. \cite{cai2021hypergraph} develop a hypergraph structure autoencoder. This model combines hypergraph structure regularization with deep autoencoders and employs a joint learning strategy.
For cross-modal image and text classification, Malik et al. \cite{Malik2023} propose a cross-modal semantic autoencoder. It utilizes three-factor non-negative matrix factorization and hypergraph regularization to map data from different sensors into a shared low-dimensional latent space.
Another approach 
incorporates both features and structure as inputs, similar to the framework outlined in (\ref{eq5ed}). For example,
Banka et al. \cite{banka2020multi} introduce a hyperconnectome autoencoder to improve brain connectivity mapping and classification. This model constructs hyperconnected groups to capture high-order relationships and employs hypergraph convolutional layers and adversarial regularization to ensure that embeddings align with the original distribution. Hu et al. \cite{hu2021adaptive} design an adaptive hypergraph autoencoder (AHGAE), which includes an adaptive hypergraph Laplacian smoothing filter. This filter reduces high-frequency noise and enhances important features by merging node and adjacency features.
Shi et al. \cite{SHI2023110172} propose a fault diagnosis method based on deep hypergraph autoencoder embeddings, which constructs unlabelled vibration signals as hypergraphs. And it designs a hypergraph convolutional extreme learning machine autoencoder to extract subspace structural information.
In social network analysis, Li et al. \cite{li2023shgae} establish a social hypergraph autoencoder, which initializes node vectors by using a hypergraph jump embedding strategy. It combines heterogeneous information propagation with attention mechanisms.
Later, Huang et al. \cite{Huang10650073} develop a self-supervised masked hypergraph autoencoder for spatio-temporal prediction. This model captures unpaired spatiotemporal relationships through a dynamic hypergraph structure and optimizes feature reconstruction. For group recommendation, Zhao et al. \cite{Zhao2024DHMAE} introduce a decoupled hypergraph masked autoencoder enhancement method (DHMAE). It combines decoupled HGNNs with graph-masked autoencoders to generate self-supervised signals and avoid reliance on high-quality data augmentation. To predict miRNA-disease associations, Xie et al. \cite{XIE2024107992} propose an enhanced hypergraph convolutional autoencoder, which expands the node receptive field through enhanced hypergraph convolutional networks.

\vspace{2mm}
These methods demonstrate the innovation of HGAEs.
However, there are several challenges:
i) HGAEs achieve information compression by mapping high-dimensional representations into a lower-dimensional latent space  and reconstructing the hypergraph structure. 
To fully reconstruct the original hypergraph data,
learning high-quality latent representations with minimal information loss is necessary \cite{SHI2023110172}.
ii) HGAEs 
produce only a single latent representation, which ignores the diversity and ambiguity inherent in data. 
When data exhibits potential uncertainty or contains noise, it becomes essential to model the latent distribution. As discussed in the subsection \ref{se351}, variational hypergraph autoencoders provide a powerful alternative for this issue.


\subsection{Hypergraph Recurrent Networks}


Hypergraph recurrent networks (HGRNs) are a cutting-edge neural network architecture specifically designed to handle temporal hypergraph data.
Similar to graph recurrent networks \cite{2018Seo,s23177534}, they utilize the recurrent neural modules (such as LSTM or GRU) to process the temporal characteristics of the hypergraph sequences and perform time-series prediction. 
In practice, HGRNs can be applied across various fields, such as predicting traffic flow \cite{2020HGC-RNN}, network intrusion detection \cite{2024HRNN}, and stock prediction \cite{2022DyHCN,2024CHNN}.

For a time-series prediction problem  on a hypergraph sequence $\mathcal{G}^{(\mathrm{t})}=(\mathcal{V}^{(\mathrm{t})}, \mathcal{E}^{(\mathrm{t})})$, where $\mathrm{t} = 1, \ldots, \mathrm{T} $.
A time-series attribute matrix $\mathbf{F} \in \mathbb{R}^{n \times \mathrm{T}}$ is defined, where $n$ represents the number of nodes, and $\mathrm{T}$ represents the number of node attributes.
Let $\mathbf{F}^\mathrm{t}$ represent the hypergraph signal observed at time $\mathrm{t}$, a common goal is to learn the mapping function $f$ from $\mathrm{T}'$ historical hypergraph signals to  $\mathrm{T}$ future hypergraph signals:
\begin{equation}
\left[\mathbf{F}^{\mathrm{t}-\mathrm{T}'+1},\cdots, \mathbf{F}^{\mathrm{t}}; \mathcal{G}\right] \xrightarrow{f(\cdot)}
\left[\mathbf{F}^{\mathrm{t}+1},\cdots, \mathbf{F}^{\mathrm{t}+\mathrm{T}}; \mathcal{G}\right]
\end{equation}

As an early exploration, Yi et al. \cite{2020HGC-RNN} first combine hypergraph convolution with recurrent neural networks and design the hypergraph convolutional recurrent neural network (HGC-RNN). It not only captures the structural dependency of data through hypergraph convolution operations but also learns the temporal dependency of data sequences by using recurrent neural layers. 
Later, 
Yin et al. \cite{2022DyHCN} propose a dynamic hypergraph convolutional network (DyHCN).
It effectively captures spatiotemporal relationships through a recurrent neural network layer, dynamic hypergraph construction, spatiotemporal hypergraph convolution, and collaborative prediction modules.
Following this, Liang et al. \cite{2023MixRNN+} design a mixed-order relation-aware recurrent neural network (MixRNN+), which captures mixed-order spatial relationships between nodes through the Mixer building block. And it adopts a new residual update strategy to handle underlying physical laws and achieve more accurate prediction of spatiotemporal sequences. Further, Yin et al. \cite{2024CHNN}  propose a collaborative hypergraph neural network (CHNN), which learns complex spatiotemporal relationships through collaborative hypergraph message passing modules.
It employs the multi-view gated recurrent neural network with an attention mechanism to achieve long-term aggregation.
Chen et al. \cite{2023HGRNN} propose a hypergraph recurrent neural network model, which introduces a semi-principled hypergraph construction method to address the challenge of missing hypergraph information. And it integrates temporal components in the encoder-decoder framework to enhance learning from temporal locality. In network traffic prediction, Yu et al. \cite{2023RHCRN} develop a routing hypergraph convolutional recurrent network (RHCRN) by regarding the routing path as a hyperedge. It uses hypergraphs to model the connections between network nodes and combines hypergraph convolutional networks with gated recurrent units to extract spatiotemporal correlations. For knowledge graph reasoning, Guo et al. \cite{2023RE-H2AN} develop a hierarchical
hypergraph recurrent attention network (RE-H$^2$AN), which performs multi-level modeling on type-induced entity hypergraphs. This model captures evolutionary patterns at different semantic granularities and achieves reasoning of high-order interactions between entities in temporal knowledge graphs. Subsequently, Tang et al.  \cite{2024DHper} propose a recurrent dual dypergraph neural network (DHyper) model to capture high-order correlations between entities and relationships simultaneously. It achieves high performance in event prediction of temporal knowledge graphs through dual hypergraph learning modules and dual hypergraph message passing networks. In network intrusion detection, Yang et al. \cite{2024HRNN} propose the hypergraph recurrent neural network, which enrichly represents traffic data through regarding each flow as a node. And it uses recurrent network modules to extract temporal features of traffic and effectively integrates rich spatiotemporal semantic representations.

\vspace{2mm}
These studies have shown the effectiveness of HGRNs in handling temporal hypergraph data. 
However, there are some challenges: 
i) In practical applications, balancing the spatial and temporal dependencies in a hypergraph is a major problem \cite{10314020}. Over-reliance on any single dimension may result in information loss.
ii) How to address the long-term dependency problem, which has not been sufficiently explored.  In traffic flow prediction, the traffic conditions during the morning peak may impact the predictions for the evening peak. For example, CHNN \cite{2024CHNN} incorporates a long-term temporal summarization module into the model to address this issue.


%

\subsection{Deep Hypergraph Generative Models}

Deep hypergraph generative models (DHGGMs) are sophisticated probabilistic unsupervised models 
engineered
to generate more realistic hypergraphs.  
In this subsection, we review three representative DHGGMs: variational hypergraph autoencoders (VHGAEs), hypergraph generative adversarial networks (HGGANs), and hypergraph generative diffusion models (HGGDMs).
These models can be applied in various fields, including network analysis \cite{2021Heterogeneous,LU2023109818,gailhard2024hygene}, recommendation systems \cite{Li2023},
bioinformatics \cite{Su2024Inferring,10.1093/bib/bbae274},
and medical image processing for disease diagnosis \cite{Pan2021HGGAN,zuo2021multimodal,Bi2022,Pan2024DecGAN}.

%


\subsubsection{Variational  Hypergraph Autoencoders}\label{se351}

Variational hypergraph autoencoders (VHGAEs) extend variational graph autoencoders (VGAEs) \cite{kipf2016variational,GraphVAE2018,8970775,DING202125} to hypergraphs. 
Like HGAEs, VHGAEs have a similar encoder-decoder framework. The difference is that VHGAEs are probabilistic models that combine variational Bayesian methods. 
Based on the assumption that each node follows an independent normal distribution, the encoder of VHGAEs learns the variational distribution of the latent representations with the mean $\boldsymbol{\mu}$ and variance $\boldsymbol{\sigma}^2$:
\begin{equation}
\label{eq6en}
\begin{aligned}
q (\mathbf{Z}\mid \mathbf{H},\mathbf{X}) = \prod_{i=1}^{n} q (\mathbf{z}_i\mid \mathbf{X},\mathbf{H})
\end{aligned}
\end{equation}
where $q (\mathbf{z}_i\mid \mathbf{X},\mathbf{H})=\mathcal{N}(\boldsymbol{\mu}_i, \text{diag}(\boldsymbol{\sigma}_i^2))$ represents the Gaussian distribution. It is conditioned on the feature matrix $\mathbf{X}$ and the incidence matrix $\mathbf{H}$. Its parameters $\boldsymbol{\mu}=\text{HGCN}_{\boldsymbol{\mu}}(\mathbf{X},\mathbf{H})$ and $\boldsymbol{\sigma}^2=\text{exp}(\text{HGCN}_{\boldsymbol{\sigma}^2}(\mathbf{X},\mathbf{H}))$
are obtained from  two HGCN modules. Then, the node embeddings are randomly sampled from the variational distribution, and stacked to obtain $\mathbf{Z}$. Finally, the decoder $p(\mathbf{H} \mid \mathbf{Z})$ reconstructs the structure  by using $\sigma(\mathbf{Z}\mathbf{Z}^{\top})$, which is similar to HGAEs.
%
The model's optimization objective is to maximize the Evidence Lower Bound (ELBO), which can be expressed as:
\begin{equation}
\mathcal{L}_{\text{ELBO}} = \mathbb{E}_{q(\mathbf{Z}\mid \mathbf{H},\mathbf{X})}[\log p(\mathbf{H} \mid \mathbf{Z})] - \text{KL}(q(\mathbf{Z}\mid \mathbf{H},\mathbf{X}) \| p(\mathbf{Z}))
\end{equation}
Here, $\mathbb{E}_{q(\mathbf{Z}\mid \mathbf{H},\mathbf{X})}[\log p(\mathbf{H} \mid \mathbf{Z})]$  is the expected log likelihood with higher values indicating better reconstruction. 
$\text{KL}(q \| p )$ is the KL divergence 
to ensure that the learned latent distribution aligns with the prior distribution $p(\mathbf{Z})$.

In early research, 
VHGAEs integrate hypergraph constraints into the latent space of conventional VAEs for model training.
For instance, Kajino \cite{kajino2019molecular} designs a molecular hypergraph grammar variational autoencoder (MHG-VAE). 
This model can generate valid molecules that conform to chemical constraints and simplify the model architecture for easier training and optimization. Building on MHG-VAE, Koge et al. \cite{koge2021embedding} improve the model by introducing metric learning. It aligns the distance similarity in the embedding space with the label space to address the consistency issue in physical property representation.
Later, researchers incorporate the hypergraph structure into the encoder's input, similar to (\ref{eq6en}).
Liu et al. \cite{liu2022hypergraph} propose a hypergraph variational autoencoder model to mine high-order interactions in multimodal data. It adopts mask-based variational inference to enhance feature representation. For efficiently processing multimodal data, Mei et al. \cite{MEI2024128225} propose a new model by combining variational autoencoders, hypergraph convolutional networks, and hyperbolic geometry. The encoder leverages neighborhood aggregation and nonlinear activation to enhance data structure understanding.
For link prediction, Fan et al. \cite{2021Heterogeneous} develop the heterogeneous hypergraph variational autoencoder (HeteHG-VAE). It maps heterogeneous information networks to hypergraphs and learns deep latent representations of nodes and hyperedges through the VAE framework. Next, Lu et al. \cite{LU2023109818} introduce a neighborhood overlap-aware heterogeneous hypergraph neural network (NOH) for link prediction. It captures high-order heterogeneity and combines neighborhood overlap-aware graph neural network to generate node representations with structural information. In the area of fault diagnosis and domain adaptation, Luo et al. \cite{10535302} establish the multi-channel variational hypergraph autoencoder (MC-VHAE). This model separates high-frequency and low-frequency components by using discrete wavelet transforms. It also introduces multi-order neighborhood hypergraph convolutional layers to aggregate high-order feature information.
For gene regulatory networks,
Su et al. \cite{Su2024Inferring} design the hypergraph variational autoencoder model. It describes single-cell RNA sequencing data through hypergraphs and utilizes structural equation modeling to address cellular heterogeneity.
For multimodal dialogue sentiment recognition tasks, Yi et al. \cite{yi2024multiencoder} propose a framework that dynamically adjusts hypergraph connections. It effectively models complex dialogue relationships through the variational hypergraph autoencoder and contrastive learning. And it alleviates contextual redundancy and over-smoothing issues in multimodal sentiment recognition.
Xia et al. \cite{10.1093/bib/bbae274} introduce a protein complex prediction model called HyperGraphComplex. This model integrates higher-order topological structures and protein sequence features from PPI networks using the variational hypergraph autoencoder. It trains both the encoder and decoder simultaneously to generate latent feature vectors for protein complexes.

\subsubsection{Hypergraph Generative Adversarial Networks}

Hpergraph generative adversarial networks (HGGANs) are an extension of graph generative adversarial networks (Graph GANs) \cite{GraphGAN2018,Ma_2019_CVPR,BESSADOK2021102090,sasagawa2021recommendation}. 
They 
simulate 
hypergraphs
through the adversarial process of competition between the generator and discriminator models. The discriminator measures the probability that the generated hypergraph comes from the real hypergraph.
Similar to Graph GANs, the optimization of the generator $G_\theta(\mathbf{Z})$  and discriminator $D_\phi(\mathbf{H},\mathbf{X})$  in HGGANs can be expressed as the following two-player minmax game:
\begin{equation}
\label{eq7}
\min_\theta \max_\phi \,\, \mathbb{E}_{\mathbf{H},\mathbf{X} \sim P_{\text{data}}}[\log {D}_\phi(\mathbf{H},\mathbf{X})] + \mathbb{E}_{\mathbf{Z} \sim P{(\mathbf{Z})}}[\log(1 - D_\phi(G_\theta(\mathbf{Z})))]
\end{equation}
where
\(P_{\text{data}}\) is the true data distribution of the given hypergraph,  
and $P{(\mathbf{Z})}=\mathcal{N}(\mathbf{0},\mathbf{I})$ is the random noise distribution. 


In recent years, Pan et al. \cite{Pan2021HGGAN} first design a novel 
HGGAN model for Alzheimer's disease analysis. 
This model generates multimodal (such as fMRI and DTI) connectivity of brain networks  through an interactive hyperedge neuron module. 
Meanwhile, Zuo et al. \cite{zuo2021multimodal} propose an innovative multimodal representation learning and adversarial hypergraph fusion (MRL-AHF)
framework for diagnosing Alzheimer's disease. It combines adversarial strategies with pre-trained models to address fusion issues caused by small multimodal data scales and inconsistent distributions between modalities. Bi et al. \cite{Bi2022IHGC-GAN} construct an influence hypergraph convolutional generative adversarial network (IHGC-GAN) model for disease risk prediction. It updates and propagates lesion information between nodes using graph convolutional networks (GCNs) and enhances predictive capability through adversarial training of the generator and discriminator. Building on this, Bi et al. \cite{Bi2022} further improve it by the hypergraph
structural information aggregation generative adversarial networks (HSIA-GANs) to aggregate structural information. This model not only achieves automatic classification of Alzheimer's samples but also accurately extracts key features to enhance the accuracy of disease classification. Zuo et al. \cite{zuo2024} use a prior anatomical knowledge-guided adversarial learning and hypergraph-aware network (PALH) to integrate multimodal medical imaging data and predict abnormal brain connectivity during the progression of Alzheimer’s disease. Pan et al. \cite{Pan2024DecGAN} establish a decoupled generative adversarial network (DecGAN) that decouples brain networks to extract neural circuits associated with Alzheimer's disease. It enhances the detection accuracy of abnormal neural circuits and the robustness of the model by combining hypergraph analysis modules and sparse capacity loss.
In addition, hypergraph generative adversarial networks have also been applied in recommendation systems. For instance, Li et al. \cite{Li2023} apply hypergraph neural networks and decoupled representation learning in recommendation systems to explore the complex relationships among items within and across baskets. The basket-level recommendations are optimized  through adversarial networks, thus improving the performance and diversity of the next-generation basket recommendation systems. 

\subsubsection{Hypergraph Generative Diffusion Models}

Hypergraph Laplacian is pivotal not only in hypergraph spectral clustering and spectral HGCNs,  
but also serves as a discrete Laplacian operator in hypergraph diffusion. This operator facilitates modeling the diffusion and propagation of 
dynamic processes in networks.
Similar to the graph Laplacian diffusion model \cite{abdelnour2018functional}, 
the diffusion process on a hypergraph can be formulated as:
\begin{equation}
\label{diff}
\frac{d \phi(t)}{d t} = -\alpha \mathbf{L} \phi(t)
\end{equation}
where $\phi(t)$ represents node signals that vary over time $t$, and $\alpha$ is a diffusion rate coefficient. 
The Laplacian operator $-\mathbf{L}$ reflects how information spreads across the hypergraph. 
Process (\ref{diff})
is often applied to tasks such as semi-supervised learning and information diffusion \cite{zhou2006learning,chitra2019random}.

In semi-supervised learning tasks, this diffusion process  enables information from labeled nodes to propagate and guide the classification of unlabeled nodes. 
For instance, Zhang et al.  \cite{Zhang2020} introduce a semi-supervised learning model on directed hypergraphs and utilize confidence intervals to address the non-uniqueness of optimal solutions. 
Prokopchik et al. \cite{Prokopchik2022} propose the nonlinear hypergraph diffusion (HyperND) model for node classification, which propagates both node features and labels and demonstrates the global convergence. Aviles et al. \cite{Aviles2022} develop a semi-supervised hypergraph learning model for Alzheimer's diagnosis by using a dual embedding strategy to capture higher-order relationships between multimodal data. For node classification, Wang et al. \cite{Wang2023} introduce the equivariant diffusion-based HNN model named ED-HNN.  It combines star expansion of hypergraphs and effectively approximates any continuous equivariant hypergraph diffusion operator. 
In information diffusion tasks, this process simulates how information spreads through a hypergraph and gradually converges to a stable state. Antelmi et al. \cite{Antelmi2020} design a method to solve the social influence diffusion problem on hypergraphs. It identifies a minimal subset of influential nodes by using a greedy heuristic. Fountoulakis et al. \cite{Fountoulakis2021} propose a local diffusion method based on a primal-dual formulation, which offers Cheeger-type guarantees for local hypergraph clustering independent of edge size. 
Sun et al. \cite{Sun2022} develop the memory-enhanced sequential hypergraph attention networks (MS-HGAT) that integrates global social dependencies and temporal information. Meanwhile, %
Jin et al. \cite{Jin2022} propose dual-channel hypergraph neural networks by capturing the users' dynamic interactions and modeling cross-content diffusion.
In neuroscience, 
Ma et al. \cite{MA2021239} incorporate positive and negative correlation links to simulate brain resting-state functional connectivity (FC). Subsequently, Ma et al. \cite{math9182345} introduce a hypergraph p-Laplacian model, which better captures nonlinear relationships between brain structural connectivity (SC) and FC. Ma et al. \cite{9965288} later improves the performance in predicting FC by introducing a sign matrix to simulate negative correlation links.

In hypergraph generative diffusion models (HGGDMs), the process of generating hypergraphs can be seen as the reverse of the diffusion process. The forward diffusion process gradually adds noise to the real hypergraph. 
During the reverse generation process, the model uses structural constraints, such as the hypergraph Laplacian, to progressively remove noise and restore the original topology.
Their development is inspired by
Markov Chain Monte Carlo techniques and stochastic process theory \cite{Dickstein2015,song2020score}. Classic denoising generative diffusion models 
typically add noise to data as a forward diffusion process, and then remove the noise gradually through a reverse diffusion process to generate data \cite{NEURIPS2020_4c5bcfec,song2020denoising,NEURIPS2023}. 
The forward diffusion process is generally designed as a stochastic process based on Gaussian noise 
\cite{NEURIPS2020_4c5bcfec}:  
\begin{equation}
\label{ddiff}
x_{t+1} = \sqrt{\alpha_{t+1}} x_t + \sqrt{1 - \alpha_{t+1}} \epsilon
\end{equation}
where $ \alpha_{t+1} $ is an element of the pre-determined noise variance sequence. 
And $ \epsilon $ is random noise sampled from the standard normal distribution $ \mathcal{N}(0, 1) $. 
As the first attempt to apply this idea of denoising generative diffusion for hypergraphs, Gailhard et al.\cite{gailhard2024hygene} propose a hypergraph generative diffusion model called HYGENE. 
Its noise diffusion and injection mechanism also satisfy the recursive formula (\ref{ddiff}).
In the generation process, HYGENE is trained through a denoising process to progressively construct complex global structures and local details.
This model offers a new solution for the generative task of hypergraphs.

\subsubsection{Open Problems}

DHGGMs reconstruct hypergraphs  in a probabilistic manner through variational inference (as in VHGAEs), generative adversarial training (as in HGGANs), or reverse diffusion-based process (as in HGGDMs).
Each approach has 
its own strengths and weaknesses:
i) VHGAEs offer stable training and continuous, interpretable latent spaces.  
However, they rely on the node independence assumption, which limits their ability to capture complex hypergraph structures. And
overly strong KL divergence forces the model to constrain the latent space too tightly, 
which limits the model's ability to capture detailed patterns in the data.
ii) HGGANs typically generate higher-quality and more realistic hypergraphs compared to VHGAEs.  
However, HGGANs often face unstable training due to some issues like mode collapse and vanishing gradients. Additionally, their latent spaces lack interpretability and continuity. 
iii) HGGDMs offer strong generation stability, especially for noisy data and tasks that strongly demand structural coherence \cite{gailhard2024hygene}. However, their step-by-step generation strategy leads to lower efficiency, 
which may limit their practical application.
%
iv) In certain fields, these models encounter some common challenges. 
For example, in tasks like molecular generation and drug discovery,
they must generate hypergraphs with specific constraints like chemical validity and synthetic feasibility \cite{koge2021embedding}. These constraints require models to not only produce high-quality structures but also ensure their practicality.

\section{Conclusion and Outlook}\label{sec4}

In the past few years, a variety of HGNNs and their variants have been developed.
As emerging deep learning models, HGNNs extend traditional GNNs to hypergraphs, which enables them to handle data with higher-order relationships. 
This paper has provided a comprehensive overview of recent advances and applications within this field. We first introduce the definitions of hypergraphs and tasks of HGNNs.  Subsequently, we categorize the mainstream  models 
into five types: HGCNs, HGATs, HGAEs, HGRNs, and DHGGMs. HGCNs are further divided into spectral HGCNs and spatial HGCNs, while DHGGMs include VHGAEs, HGGANs, and HGGDMs. Finally, we review the current research for each category, which covers application scopes, underlying mechanisms, contributions, and open problems. Through the exploration of various architectures, we have witnessed 
that HGNNs play an increasingly critical role in  processing complex hypergraph data.



While existing HGNNs have demonstrated remarkable capabilities across various domains, challenges still persist and require further exploration and resolution:
i) 
The challenge of hypergraph construction lies in precise mathematical modeling tailored to specific tasks.
There is no standardized method to construct hypergraphs. 
In different application scenarios, 
the definitions of nodes and hyperedges vary significantly, such as explicitly constructing hypergraphs from structural information or implicitly constructing hypergraphs from feature information \cite{10040228,10068184,YIN2024114586}.  It is crucial to design appropriate construction techniques tailored to the specific problem at hand.
ii) 
Existing HGNNs generally perform well on small-scale datasets, but their scalability on large-scale datasets has not been fully validated. To handle large-scale data, these models need to be capable of maintaining accuracy while reducing computational and storage costs.
Current approaches include factorization-based hypergraph reduction and hierarchical hypergraph learning \cite{9667788,Dai2023}. It is essential to develop new methods to enhance the efficiency and performance of large-scale hypergraph learning.
iii) 
Explainability research on HGNNs seeks to uncover the internal mechanisms, 
such as identifying the most important nodes, hyperedges, or sub-hypergraphs, and understanding how patterns of information interaction and aggregation affect performance. For instance,  HyperEX \cite{maleki2023learning} is a new post-hoc explainability method for HGNNs. However,
explainability research on HGNNs is still in its early stages, with an urgent need for standardized frameworks to enhance model transparency and reliability in practical applications.
iv) Heterogeneous hypergraphs are common in real-world applications, 
such as recommendation systems, social networks, and knowledge graphs. A key challenge is effectively aggregating information from different node types through heterogeneous hyperedges. To address this, strategies like attention-based meta-path aggregation \cite{2023Hypergraph} and neighborhood overlap-aware aggregation \cite{LU2023109818} have been proposed. How to improve models for better information aggregation in heterogeneous hypergraphs remains a key research focus. 

As research progresses,
further breakthroughs and innovations in HGNNs are expected to
provide more effective solutions for tackling complex real-world challenges.

\backmatter

\bmhead{Authors Contributions}
The authors make their contributions equally.

\bmhead{Conflict of Interest}
The authors declare 
no conflict of interest. 
\bibliographystyle{plain}
\bibliography{ref}

%
%
%
%
%
%
%
%
\end{document}